\documentclass{article}
\usepackage{ijcai21}

\usepackage{times}

\usepackage{soul}
\usepackage{url}
\usepackage[hidelinks]{hyperref}
\usepackage[utf8]{inputenc}
\usepackage[small]{caption}
\usepackage{graphicx}
\usepackage{amsmath}
\usepackage{booktabs}
\title{CoRe: An Efficient Coarse-refined Training Framework for BERT}

\author{
Cheng Yang \footnote{Equivalent contribution}\and
Shengnan Wang $ ^*$ \and
Yuechuan Li \and
Chao Yang \and
Ming Yan\and
Jingqiao Zhang\and
Fangquan Lin \\
\affiliations
Alibaba Group \\
\emails
\{charis.yangc, shengnan.wsn, yuechuan.lyc, xiuxin.yc, ym119608,  jingqiao.zhang, 	fangquan.linfq\}@alibaba-inc.com
}

\begin{document}

\maketitle
\begin{abstract}
	In recent years, BERT has made significant breakthroughs on many natural language processing tasks and attracted great attentions. Despite its accuracy gains, the BERT model generally involves a huge number of parameters and needs to be trained on massive datasets, so training such a model is computationally very challenging and time-consuming. Hence, training efficiency should be a critical issue. In this paper, we propose a novel coarse-refined training framework named CoRe to speed up the training of BERT. Specifically, we decompose the training process of BERT into two phases. In the first phase, by introducing fast attention mechanism and decomposing the large parameters in the feed-forward network sub-layer, we construct a relaxed BERT model which has much less parameters and much lower model complexity than the original BERT, so the relaxed model can be quickly trained. In the second phase, we transform the trained relaxed BERT model into the original BERT and further retrain the model. Thanks to the desired initialization provided by the relaxed model, the retraining phase requires much less training steps, compared with training an original BERT model from scratch with a random initialization. Experimental results show that the proposed CoRe framework can greatly reduce the training time without reducing the performance.
\end{abstract}




\maketitle
\section{Introduction}
Large-scale pre-trained language models, such as BERT \cite{BERT}, XLNet \cite{Xlnet},  GPT \cite{GPT}, have achieved great success in Natural Language Processing (NLP).  
In recent years, a lot of BERT based deep models were proposed, such as  RoBERTa \cite{ROBERT}, ALBERT \cite{ALBERT}, Structbert \cite{structbert}, etc., most of which yielded new state-of-the-art results in many NLP tasks. In addition to NLP, the BERT structure is also applied to computer vision \cite{su2019vl},  speech recognition \cite{lin2019enhanced}, recommendation systems \cite{BERT4Rec},  and many other engineering fields. All of these BERT models follow the pre-training paradigm: pre-trained on massive unlabeled datasets and fine-tuned on a small downstream labeled dataset for a specific task.

Though significant improvements are obtained by BERT models, they usually involve over hundreds of millions of parameters and training such models is very time-consuming. For example, BERT-Base has more than 110M parameters, and training  BERT-Base takes 4 days on 4 TPUv3 \cite{BERT}. Such a drawback may restrict its practical application. Hence, training efficiency should be an important issue.

It should be mentioned that many well known model compression \cite{Deepcompre,Hansong} techniques, including  knowledge distillation \cite{distillation,Hinton}, quantization  \cite{shen2020q}, pruning \cite{gordon2020compressing,mccarley2019pruning}, can only be used for speedup in the inference stage. Namely, these methods are implemented after the model has been fully trained. In addition, to ensure performance, it is usually encouraged to first train a very large  model and then heavily compress \cite{li2020train}, which makes the  training stage even more challenging. In contrast, we focus on the speedup of the training stage, which is generally a more difficult task.

In this paper, we propose an efficient coarse-refined training framework named CoRe to accelerate BERT training.   Specifically, we decompose the training process of BERT into two phases. In the first phase, we design a relaxed model   by introducing a fast self-attention mechanism and factorizing the large parameters in the feed-forward network sub-layer. The relaxed model has a similar structure but much lower model complexity and much less parameters compared with the original standard BERT, so it can be quickly trained. In the second phase, we recover the original BERT from the trained relaxed model, and further retrain the  model to make the network better behaved. Thanks to the desired initialization provided by the relaxed model,  retraining the recovered model is much more time-efficient than training an original BERT from scratch using a random initialization. 



We provide extensive experiments on NLP benchmarks to show the effectiveness and efficiency of the proposed CoRe training method. The results demonstrate that 1) using the same training steps, the accelerated method  achieves similar performance, compared with the original method, but consumes much less training time; 2) using the same training time, the accelerated method  achieves better performance than the original method.  In addition, we show that  the proposed CoRe training framework can be combined with the other training speedup methods to further improve the training efficiency.

The rest of the paper is organized as follows. In Section 2, we briefly review the related work. In Section 3, we introduce the main architecture of BERT. In Section 4, we propose the CoRe training method. In Section 5, we present  experimental results to show the efficiency and effectiveness of the proposed method.   Finally, we draw the conclusion in Section 6.
\section{Related Work}
Bidirectional Encoder Representations from Transformers, which is called BERT \cite{BERT}, has shown its  powerful representative ability and achieved state-of-the-art results in eleven NLP tasks.  
Following BERT, many high-performance models were proposed. By using more training data and more training steps, RoBERTa \cite{ROBERT} achieved better performance than BERT.  KBERT \cite{liu2019k} injected domain knowledge into the BERT model  and outperformed BERT in many domain-specific tasks. StructBERT \cite{structbert} replaced the next sentence prediction task by a three-way classification task during pre-training and achieved  state-of-the-art results on many downstream tasks, including GLUE  and SQuAD v1.1.  ELECTRA \cite{clark2020electra} and MC-BERT \cite{xu2020mc} further introduced more effective pre-training task and achieved better performance. 

However, almost all the BERT models involve a huge number of parameters, which makes training such models both computationally and storage  challenging. Some pioneering attempts have been made to accelerate the training of BERT.

\subsection{BERT Training Speedup}\label{sec:2-1}
A direct way to accelerate BERT training is to increase the batch size by using more machines and train the model in a distributed manner.  However, traditional stochastic optimization methods, such as the stochastic gradient descent (SGD), perform poorly in large mini-batches training. Naively increasing the batch size generally leads to degraded performance and reduced computational benefits \cite{goyal2017accurate}. To tackle this issue, in  \cite{Largebatch}, the authors proposed a layerwise adaptive large batch optimization method, named LAMB, which is able to train the BERT model with extremely large mini-batches. By using 1024 TPUv3 chips, the training time of BERT was reduced from 3 days to 76 minutes. Though tens of times speedup is achieved, such a method requires a huge amount of computing and storage resources, which are unavailable for common users.  

ALBERT \cite{ALBERT} introduced two parameter reduction techniques, namely factorized embedding parameterization and cross-layer
parameter sharing, which significantly reduced the memory consumption. In addition, ALBERT also achieved training speedup due to less communication in the distributed training setting,  since the communication overhead mainly depends on the number of parameters. However, ALBERT has the same architecture and  almost
the same computational complexity as BERT, so the speedup is limited and  training ALBERT is still very time-consuming.


To accelerate the BERT training in an algorithmic sense, in \cite{Stacking}, the authors proposed a progressively stacking method, which trains a deep network by repeatedly stacking from a shallow one. This method is based on the observation that different layers in a trained BERT model have similar attention distributions. Utilizing the natural similarity characteristic, progressively stacking achieved training speedup without performance degradation.


%
%


\subsection*{Efficient Attention Mechanisms}
In the literature, there are also some researches focusing on improving the computational efficiency of self-attention, which is the main component of BERT. 
%
The standard content-based attention mechanism is usually computationally expensive especially when the input sequence is long. 
In \cite{line_att}, the authors proposed an efficient attention mechanism, which has substantially less memory and computational costs   than the traditional  dot-product attention. In \cite{wang2019less}, the authors modified the self-attention module in the generative adversarial networks, which also improves the network training efficiency. In \cite{katharopoulos2020transformers}, the authors expressed  self-attention as a linear dot-product of kernel feature maps, which reduces the complexity of self-attention from $\mathcal{O}(n^2)$ to $\mathcal{O}(n)$, where $n$ is the input sequence length.
However, most of these modifications  dramatically change  the original structure of the  attention module, which may make the attention layer less effective. 
In addition,   these methods were only demonstrated to be useful in the field of computer vision. It is still unclear whether they can be used in the BERT models for dealing with the NLP tasks. 



\subsection{Similarity Matrix Approximation} \label{sec:2-3}
In this paper, we propose a fast self-attention mechanism  for the BERT model.  Some techniques in our method is inspired by the anchor graph hashing method \cite{liu2011hashing}. Graph hashing is an efficient method used for retrieval \cite{weiss2009spectral}. Generally, it needs to construct a large graph matrix to store the similarity information among the data, which is very similar to the query-key similarity matrix in the self-attention mechanism.  Directly computing the graph matrix is inefficient in terms of both computation and storage. Anchor graph hashing addresses this problem by selecting  some anchor points from the data and utilizing the anchor points to approximate the graph matrix. Similar idea can also be used to accelerate the computing of the query-key similarity matrix self-attention mechanism. For more details about anchor graph hashing, one can refer to \cite{liu2011hashing,wang2018distributed}.

\section{Introduction of BERT and Complexity Analyses}
\begin{figure}[t]
	\centering
	\includegraphics[scale=0.5]{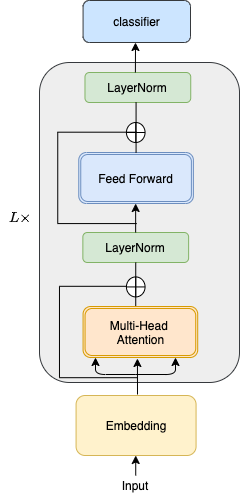}
	\caption{\label{fig:bert} The BERT model  architecture.}
\end{figure}

In this section, we give a brief introduction of BERT,  and analyze the computational complexity of the its components.  In general, the BERT model consists of an embedding layer (input layer), a classifier (output layer), and multiple encoder layers. Each encoder layer involves a multi-head self-attention sub-layer and a feed forward network (FFN) sub-layer.  The architecture of BERT is shown in Figure \ref{fig:bert}.

\subsection{Multi-head Self-attention} \label{sec:3.1}
Multi-head self-attention module is a key component of the BERT encoder layer.  A self-attention function can be formulated as
\begin{equation}\label{eq:1}
\begin{split}
&Att(Q,K,V) =\text{softmax}(\frac{QK^T}{\sqrt{d_k}})V,\\
&\text{where} \quad Q\in \mathrm{R}^{n\times d_k}, K\in \mathrm{R}^{n\times d_k}, V \in\mathrm{R}^{n\times d_v}.
\end{split}
\end{equation}
Here, we call $Q=[q_1,q_2,\dots, q_n]^T, K=[k_1,k_2,\dots, k_n]^T$, and $V=[v_1,v_2,\dots, v_n]^T$ the query, key,  value matrix, respectively, and $q \in \mathrm{R}^{d_k}, k\in\mathrm{R}^{d_k}, v\in\mathrm{R}^{d_v}$  denote a query, key, value vector, respectively. In the BERT model, we set $d_k=d_v=d'=d/h$ \cite{BERT}, where $d$ is the hidden size.
We call the matrix
\begin{equation} \label{eq:5}
S=\text{softmax}(QK^T/\sqrt{d_k})
\end{equation}
the query-key similarity matrix, which is a stochastic matrix, i.e. the entries are all non-negative and the sum of each row is 1.
We call each row of $S$ the similarity vector, which defines the relative similarity between a query and all the keys.
From (\ref{eq:1}), we can see that the output of self-attention for a query vector $q$ (a row of $Q$)  is a weight average of the value vectors (rows of $V$) with the corresponding row of $S$ as the coefficient.

The multi-head self-attention sub-layer  involves $h$ independent self-attention modules. Let $X\in \mathrm{R}^{n\times d}$ denote the input sequence of the encoder layer. The output of  multi-head self-attention  is formulated as
\begin{equation} \label{eq:2}
MulHeadAtt(X) = Concat(Z_1, Z_2, \dots, Z_h)W^O,
\end{equation}
where $Z_i=Att(Q_i, K_i,  V_i)$ is the output of the $i$-th head. Here $Q_i=XW_i^Q ,K_i=XW_i^K, V_i=XW_i^V$ and $W_i^Q,W_i^K \in\mathrm{R}^{d\times d_k}$,  $W_i^V \in  \mathrm{R}^{d\times d_v}$, $W^O\in \mathrm{R}^{d\times d}$.   

\subsection{Position-wise Feed-forward Network}
The FFN sub-layer includes two linear transformations with a ReLU activation in between, which can be formulated by the following function
\begin{equation}\label{eq:3}
\begin{split}
&FFN(x)=max(0, xW_1+b_1)W_2+b_2,\\
&\text{where} \quad W_1 \in \mathrm{R}^{d\times d_f}, W_2 \in \mathrm{R}^{d_f \times d}.
\end{split}
\end{equation}
The feed-forward network  is applied to each position separately and identically.  In the literature, most BERT models set $d_f = 4d$ \cite{BERT,ALBERT}.
\subsection{Add-normalization}
In the BERT model structure, both multi-head self-attention sub-layer and FFN sub-layer are followed by an add-norm operation, which is a residual connection \cite{he2016deep} with a layer normalization \cite{ba2016layer}, defined as
\begin{equation}
AddNorm(x) = LayerNorm(x+Sublayer(x)),
\end{equation}
where $x$ is the input of the sub-layer and $Sublayer(x)$ is the output.

\subsection{Computational Complexity}
Here we give the computational complexity analysis of  BERT.  The computation of BERT mainly comes from the encoder layers. In the multi-head self-attention sub-layer, the computational complexity of self-attention operation (Equation (\ref{eq:1})) is $\mathcal{O}(n^2d)$ and the computational complexity  of the multi-head projection operation (Equation (\ref{eq:2})) is $\mathcal{O}(nd^2)$. In the FFN sub-layer, the computational complexity is  $\mathcal{O}(nd^2)$

\section{Methodology}
In this section, we propose a two-phase coarse-refined training framework named CoRe, to speed up the training process of BERT. 

\subsection{Coarse Train: Pre-training  Relaxed BERT} 
As stated above, the computation of BERT mainly comes from the encoder layer, specifically, the multi-head self-attention sub-layer and FFN sub-layer. In the following, we devise a relaxed BERT model, which has similar structure but much lower model complexity, compared with the original BERT.  Speicifically, we introduce a fast self-attention  mechanism in the multi-head self-attention sub-layer and apply parameter factorization to the FFN sub-layer to simplify  the two computationally expensive modules.


\subsubsection{Fast Multi-head Self-attention Mechanism}\label{sec:4-1-1}
$\\$
According to equation (\ref{eq:1}), we see that for each query vector $q\in\mathrm{R}^{d_k}$,  self-attention mechanism computes the similarity vector of $q$ and all the keys in $K$ by
\begin{equation}\label{eq:4}
\begin{split}
s(q,K) =\text{softmax}(\frac{q^TK^T}{\sqrt{d_k}})\\
\end{split}
\end{equation}
as the weights on the values.  Since there are totally $n^2$ query-key pairs, such a step is a heavy task when $n$ is large.  We hope to compute the query-key similarities in a more efficient way. Note that if $q_1$ and $q_2$ are close, namely $||q_1-q_2||$ is very small,   the dot-products $q_1^T k$ and $q_2^T k$ should also be close, for any key $k$ from $K$. Hence, the similarity between $q_1$ and $K$ can be approximated by the similarity between $q_2$ and $K$.
Inspired by the idea of anchor graph hashing method shown in Section \ref*{sec:2-3}, for a query sequence $Q=[q_1,q_2,\dots, q_n]^T$ with length $n$, we also consider to  select $m  (m\ll n)$ anchor queries $A=[a_1, a_2, \dots, a_m]$ from $Q$, and then utilize them to approximate the query-key similarities. 
%
Before giving the method, we normalize the query vectors $q \in Q$ and anchor vectors $a \in A$ to unit length, namely ${q}=q/||q||$ and ${a}=a/||a||$.  Then large dot product $q^Ta$ implies small distance $||q-a||$. Instead of directly computing the query-key similarity matrix, 
we first compute the query-anchor similarity by
\begin{equation}\label{eq:14}
S_1(Q,A)=\text{softmax}({QA^T}*\sqrt{d_k}),
\end{equation}
and  the  anchor-key similarity by
\begin{equation}\label{eq:13}
S_2(A,K)=\text{softmax}({AK^T}),
\end{equation}
where  $S_1(Q,A)$  and $S_2(A,K)$ are matrices of size $n\times m$ and $m\times n$, respectively.
Note that we  use the scaling factor $\sqrt{d_k}$ in (\ref{eq:14}) and $1$ in  (\ref{eq:13})  instead of $1/\sqrt{d_k}$ used in the standard self-attention (\ref{eq:1}). That is because the values in ${q}$ and ${a}$ will be very small after normalization, especially when $d_k$ is large. Equation (\ref{eq:14})  uses larger  scaling factor than (\ref{eq:13}), since $K$ is not normalized.
The query-key similarity matrix is approximated by
\begin{equation}\label{eq:15}
\tilde{S}(Q,K)=S_1(Q,A)*S_2(A,K).
\end{equation}
It can be verified that $\tilde{S}(Q,K)$ is still a stochastic matrix, since both $S_1(Q,A)$ and $S_2(A,K)$ are stochastic matrices.
One can see that for any query $q_i$ from $Q$,  the query-key  similarity vector $\tilde{s}(q_i,K)$ (the $i$-the row of $\tilde{S}(Q,K)$) is computed by
\begin{equation}\label{eq:16}
\tilde{s}(q_i,K)=s_1(q_i,A)*S_2(A,K),
\end{equation}
where $s_1(q_i,A)$ is the $i$-th row of $S_1(Q,A)$, which stores the  relative similarity between the query vector $q_i$ and all the anchor vectors. 
Denote $s_1(q_i,A)$ by $s_1(q_i,A)=[\lambda_1, \lambda_2,\dots, \lambda_m]$, where $\lambda_j$ denotes the relative similarity between $q_i$ and $a_j$. 
Then equation (\ref{eq:16}) can be represented by
\begin{equation}\label{eq:6}
\tilde{s}(q_i,K)=\sum_{j=1}^{m} \lambda_j s_2(a_j,K),
\end{equation}
where $s_2(a_j,K)$ is the $j$-th row of $S_2(A,K)$, which stores the  relative similarity between the anchor vector $a_j$ and all the keys. Hence the query-key similarity vector is  approximated by a weight average of the anchor-key similarity vectors, and large weight $\lambda_j$ will be assigned if the anchor  $a_j$ is close to the query $q_i$. As stated above, the similarity between $q$ and $K$ is close to the similarity between $a$ and $K$ if $q$ and $a$ are close. Therefore, equation (\ref{eq:15}) gives a desired approximation of the query-key similarity matrix $S(Q,K)$.

\begin{figure}[t]
	\flushleft
	\centering
	\includegraphics[scale=0.3]{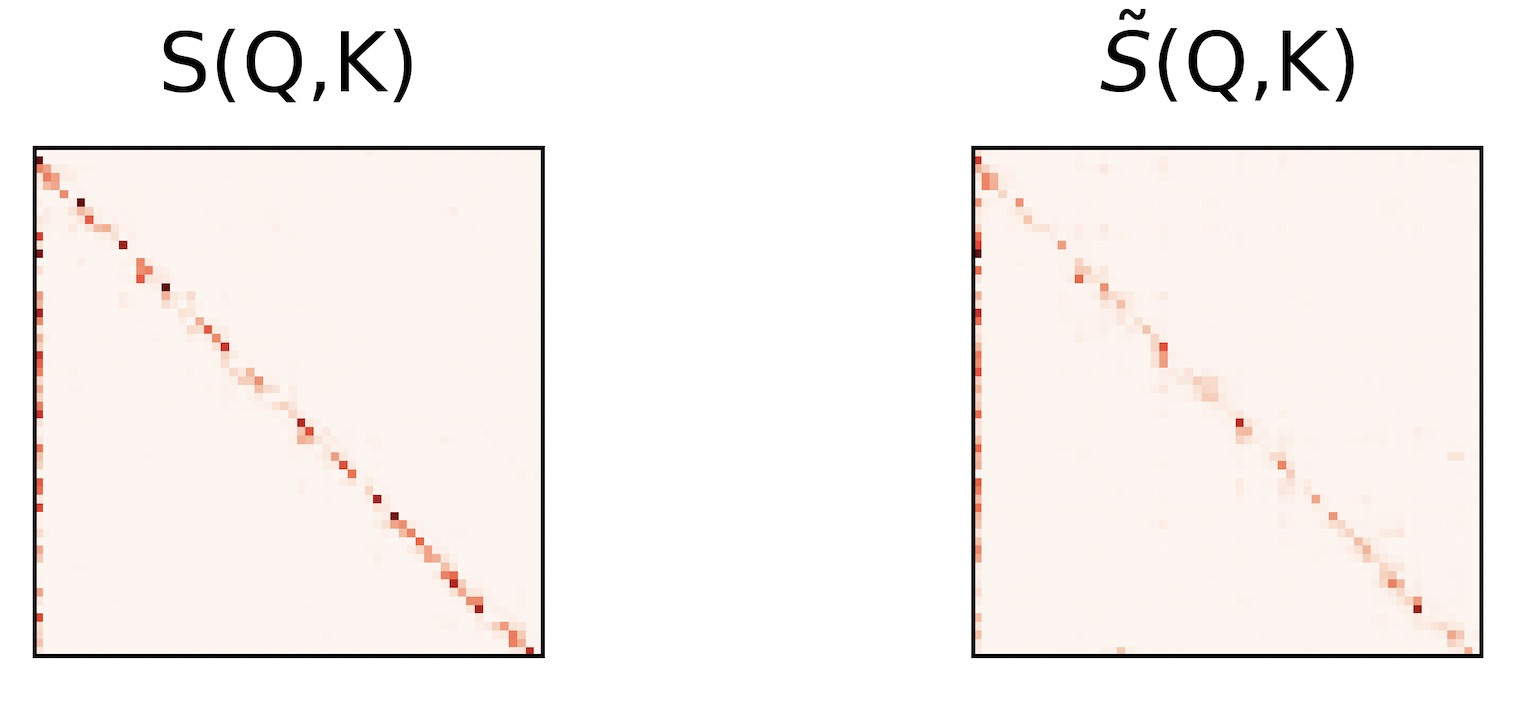}
	\caption{\label{fig:4} Visualization of $S(Q,K)$ and $\tilde{S}(Q,K)$. For a randomly chosen sample, we randomly chose an attention head and respectively compute $S(Q,K)$ by (\ref{eq:5}) and $\tilde{S}(Q,K)$ by (\ref{eq:15}), where the anchor set $A$ is randomly selected from $Q$. The figure shows that $S(Q,K)$ and $\tilde{S}(Q,K)$ are very close.}
\end{figure}

\paragraph{Anchor selection}
To make the approximation accurate,  the anchor points should be representative such that  for most query vectors $q$ in $Q$, there exists at least one neighbor anchor point  $a$ which is close to $q$. To achieve this goal, an effective and simple way is that for each given query set $Q\in \mathrm{R}^{n\times d_k}$, we can randomly select $m$ vectors from $Q$ as the anchor queries. Namely, each query set $Q$ has its own anchor set $A$. For more accurate approximation, we can first cluster the $n$ query vectors from the given $Q$, and then select representative anchor queries from all the clusters. Since the clustering step will introduce more computation, in our experiments, we adopt the random selection strategy. We randomly chose an attention head, and visualize $\tilde{S}(Q,K)$  and $S(Q,K)$ for a randomly chosen sample in   Figure \ref{fig:4}. We can see that the approximate similarity matrix $\tilde{S}(Q,K)$ is close to the real similarity matrix  $S(Q,K)$.

Based on (\ref{eq:15}), the self-attention function (\ref{eq:1}) can be rewritten as 
\begin{equation}\label{eq:9}
\begin{split}
FastAtt(Q,K,V;A)&=S_1(Q,A)S_2(A,K)V\\
&=S_1(Q,A)(S_2(A,K)V).
\end{split}
\end{equation}
By first computing  $S_2(A,K)V$, the complexity of self-attention is reduced to $\mathcal{O}(mnd)$.  Further, we also consider to reduce the complexity of the multi-head projection, namely function (\ref{eq:2}).
Denote the output of head $i$ by $Z_i = FastAtt(Q_i,K_i,V_i;A_i)$, and  denote $W^O$ in (\ref{eq:2}) by 
\begin{equation*}
W^O=[W^O_1;W^O_2; \dots; W^O_h],
\end{equation*}
where $W^O_i\in\mathrm{R}^{d'\times d}$. Then the fast multi-head self-attention  can be formulated by
\begin{equation} \label{eq:10}
\begin{split}
&FastMulHeadAtt(Q,K,V;A)\\
&\qquad=  Concat(Z_1, Z_2, \dots, Z_h)W^O\\ 
&\qquad= \sum_{i=1}^{h}Z_iW^O_i\\
&\qquad = \sum_{i=1}^{h}S_1(Q_i,A_i)(S_2(A_i,K_i)V_i)W^O_i\\
&\qquad=\sum_{i=1}^{h}S_1(Q_i,A_i)(S_2(A_i,K_i)V_iW^O_i).
\end{split}
\end{equation}
By first computing $S_2(A_i,K_i)V_iW^O_i$, the $\mathcal{O}(d^2n)$ complexity is also avoid. The fast multi-head self-attention  (\ref{eq:10}) merges the two steps, namely self-attention and multi-head projection. 
\paragraph{Complexity Comparison}
In the original multi-head self-attention mechanism shown in Section \ref{sec:3.1}, the time complexity for each head $i$ to compute the similarity matrix $S(Q_i,K_i)$  is $\mathcal{O}(n^2d)$. The time complexity  to compute the output of self-attention by $Z_i=S(Q_i,K_i)*V_i$ is $\mathcal{O}(n^2d)$. The time complexity of multi-head projection (\ref{eq:2}) is $\mathcal{O}(nd^2)$. So the total time complexity of the original multi-head self-attention mechanism  is $\mathcal{O}(n^2d+nd^2)$. 
In the proposed fast multi-head self-attention  mechanism, the time complexity for each head $i$ to compute the  similarity matrices $S_1(Q_i,A_i)$ and $S_2(A_i,K_i)$ are both $\mathcal{O}(mnd)$.  The time complexity  to compute $U_i=S_2(A_i,K_i)*V_i$ is $\mathcal{O}(mnd)$  and the time complexity to compute 
$\Phi_i=U_i*W^O_i$ is $\mathcal{O}(md^2)$. The time complexity  to compute $\Gamma_i = S_1(Q_i,A_i)*\Phi_i$ is $\mathcal{O}(mnd)$. So the total time complexity for the proposed fast   multi-head self-attention is only $\mathcal{O}(mnd+md^2)$, which is much smaller than $\mathcal{O}(n^2d+nd^2)$ since $m \ll n$. Note that  in the fast  multi-head self-attention mechanism, the computation of each step is still in parallel,  so complexity reduction implies training time reduction.




\subsubsection{Fast Position-wise FFN}
$\\$
The huge size of the parameters in the position-wise FFN sub-layers  is another reason that makes the network training time consuming. In addition, such a large number of the parameters will also take up a lot of storage resources. In  \cite{ALBERT}, the authors factorized the parameter in the embedding layer (input layer) and greatly reduced the number of parameters. However, such an operation  cannot bring significant training speed up since most computation comes from the hidden encoder layers. In the relaxed BERT model,  we also apply parameter factorization to  the FFN sub-layers.  By factorizing the $d\times d_f$ large matrix $W_1$ into two small matrices 	of size $d\times d_r (d_r \ll d)$ and $d_r\times d_f$, and factorizing the $d_f\times d$ matrix $W_2$ into matrices 	of size $d_f\times d_r$ and $d_r\times d$, we can reduce the $\mathcal{O}(nd^2)$ time complexity in (\ref{eq:3})  to $\mathcal{O}(ndd_r)$. 
In addition, the time of communication in distributed training is also reduced,  since the communication overhead mainly depends on the number of parameters. 

Nevertheless, as shown  in \cite{ALBERT}, the input layer  is only used for learning context-independent representations, while  the encoder layers  are meant to learn context-dependent representations. The powerful representational ability of BERT mainly comes from the process of learning the context-dependent representations \cite{ROBERT}. So reducing the parameters in the input layer will not significantly affect the model performance while the encoder layer's parameters reduction may lead to model degradation and hence worse performance. In addition, the above-mentioned fast self-attention mechanism may also make the model less effective, since it loses some detailed information. Such a problem can be well addressed in the retraining phase.

\subsection{Refined Training: Model Transformation and Retraining}
In the coarse training phase, we trained a relaxed BERT model, which is not directly used for dealing with specific NLP tasks.  Our goal is  still to train an original standard  BERT.  We show that the relaxed BERT model can be transformed into an original BERT model, and meanwhile the performance is inherited. Then we will retrain the model to further improve its performance.
In the following, we specifically describe how to restore the original BERT model from the trained relaxed model.

\paragraph{Embedding and classifier layers:} since the relaxed BERT model does not make any change of the Embedding (input)  and Classifier (output) layers, these parts can be directly used for the original BERT.

\paragraph{Multi-head self-attention sub-layer:} note that the weight parameters themselves in this part of the relaxed model are not changed. The proposed fast multi-head self-attention mechanism only modifies the way of computation, including introducing some approximate calculations and changing calculation order, so the parameters of this part can also   be directly used when recovering the original BERT. The only modification is that we will revert to the standard multi-head self-attention mechanism when retraining the model.

\paragraph{Feed-forward network sub-layer:} the FFN sub-layer cannot be directly  used since the parameter size does not match. In the relaxed model, we decomposed each large weight matrix of the FFN sub-layer into two relatively small weight matrices. Now we recover the large weight matrix  by  multiplying the two small matrices back. The recovered FFN sub-layer has  exactly the same functionality as FFN sub-layer of the trained relaxed model, namely it inherits the performance of the FFN sub-layer in the relaxed model. In addition, since the number of parameters increases, better performance can be achieved by further training.

%

Now we have successfully transformed the trained relaxed BERT model  into an original BERT model. Since the original BERT model has more precise attention computing and higher model complexity,  better performance can be achieved if we further retrain the model.  In addition, due to the desired initialization provided by the relaxed model, the training process is greatly shortened. Compared with directly training an original BERT model from scratch with a random initialization, our method requires much less  training time, since we can quickly reach a near-optimal state.

\subsection{Combining with The Existing Speedup Techniques}
The proposed CoRe training framework is orthogonal to most of the other speedup strategies mentioned in Section \ref{sec:2-1}, namely, it can be used in conjunction with ALBERT and progressively stacking. 

\paragraph{Combined with ALBERT} For ALBERT, the proposed CoRe framework can be directly applied. Specifically, we only need to decompose the training process of ALBERT into two phase. In the first phase, we will train an relaxed ALBERT model. The only difference between the relaxed ALBERT model. and the relaxed BERT model is that in ALBERT the parameters in the encoder layers are shared and the embedding parameter in the input layer is factorized. When the relaxed ALBERT model is trained, we can also restore an original ALBERT model using a similar way as stated above and further train the model in the second phase.

\paragraph{Combined with progressively stacking} Combined with the idea of progressively stacking, we can firstly training shallow relaxed BERT model and gradually stacking it to a deep relaxed model. When the performance of the deep relaxed BERT model does not improve any more, then we can restore a deep original BERT model as stated in last subsection and further retrain the model.

\section{Experiments}
In this section, we evaluate the proposed CoRe training method on NLP tasks. 

\subsection{Experimental Setting}

%

Following the setup in \cite{BERT}, we use English Wikipedia (2,500M words) corpus  and BookCorpus (800M words) as the datasets for pre-training the models. We set the maximum length $n$ of each input sequence to be $512$. The experiments are performed on a distributed computing cluster consisting of 64 Telsa V100 GPU cards with 32G memory, and the batch size is 64*16=1024 sequences (1024*512=524,000 tokens/batch).  We use Adam optimizer with learning rate of $2e-4, \beta_1= 0.9, \beta_2 = 0.999$, L2 weight decay of 0.01, learning rate warmup over the first 10,000 steps, and linear decay of the learning rate. All the other settings are the same as  \cite{BERT}, including the data preprocessing.

We test the proposed CoRe method on the BERT-Base model, in which the number of  hidden encoder layers is $L=12$, the hidden size is $d=768$, the number of heads is $h=12$. In the CoRe  framework, we set the number of anchors $m=32$, and the factorized size $d_r=64$. 

We denote the BERT-Base model trained using the proposed CoRe framework by $\textmd{BERT}_{fast}$ and  the  model trained using the original method \cite{BERT}  by  $\textmd{BERT}_{ori}$, which is the baseline.



\subsection{Evaluation on SQuAD}\label{sec:5-2-1}
SQuAD v1.1 is a machine reading comprehension  consisting of 100,000+ questions created by crowd workers on 536 Wikipedia articles  \cite{rajpurkar2016squad}.
We fine-tune the pre-trained models on SQuAD v1.1 for two epochs with a batch size of 12 and learning rate of 3e-5.

We report the results in terms of the training time and the F1 score   in Table \ref{tb:2}.  The F1 scores are  the median of the development set results of five random initializations.
To make a fully comparison, we show the performance  for different training steps. The total training steps differ from 100,000 to 400,000, and for fairness, in every comparison, the total training steps for $\textmd{BERT}_{ori}$ and $\textmd{BERT}_{fast}$ are same. For each  $\textmd{BERT}_{fast}$ model, we use half  steps for training the relaxed  model  at the first phase, and use the remaining steps for refined retraining. Note that we do not use enough time to train the models,   otherwise the models will not have significant differences if all of them are trained to converge.  Moreover, the training speedup has more significance when the training time and available computation resource are limited.

From Table \ref{tb:2}, we see that with the same training steps, $\textmd{BERT}_{fast}$ achieves comparable performance to $\textmd{BERT}_{ori}$, but consumes much less training time (about $25\%$ shorter).
\begin{table}[t]\centering 
	\caption{Dev set results on SQuAD v1.1. The total training steps differ from 100,000 to 400,000.}
	\label{tb:2}
	\qquad\begin{tabular}{|c|c|c|c|c|c|c|c|}
		\hline
		{Models}&Steps &Time (hours)&F1 score  \\ 
		\hline
		\hline
		$\textmd{BERT}_{ori}$&100,000&6.64& 84.2 \\ 
		\hline
		$\textmd{BERT}_{fast}$&100,000 &4.92  &84.4  \\ 
		\hline
		\hline
		$\textmd{BERT}_{ori}$&200,000&14.24&  87.5  \\ 
		\hline
		$\textmd{BERT}_{fast}$&200,000 &10.58& 87.3  \\ 
		\hline
		\hline
		$\textmd{BERT}_{ori}$&300,000&20.16& 88.1 \\ 
		\hline
		$\textmd{BERT}_{fast}$&300,000&15.21 & 88.6 \\ 
		\hline
		\hline
		$\textmd{BERT}_{ori}$&400,000&26.67& 88.2\\ 
		\hline
		$\textmd{BERT}_{fast}$&400,000&19.43 & 88.8\\ 
		\hline
	\end{tabular}
\end{table}
Note that both $\textmd{BERT}_{ori}$ and $\textmd{BERT}_{fast}$ obtain similar results to   the BERT-Base model in \cite{BERT} after training for 300,000 steps, which makes these results more convincing. 

%

\begin{figure}[t]
	\centering
	\includegraphics[scale=0.6]{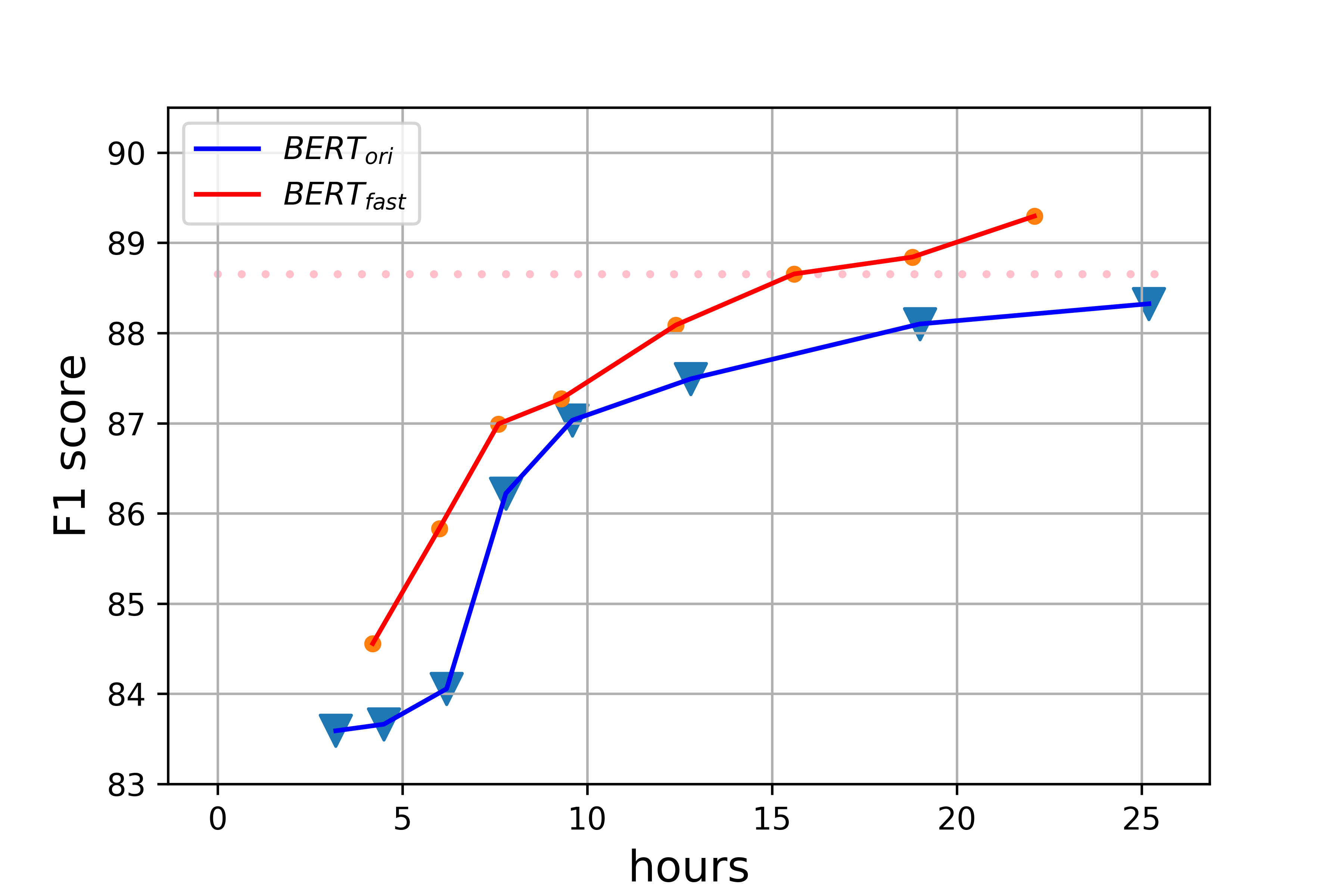}
	\caption{\label{fig:1} The F1 score curves of $\textmd{BERT}_{ori}$ and $\textmd{BERT}_{fast}$ on SQuAD v1.1. The x-axis denotes the wall time of training.}
\end{figure}

Next, we compare the performance of the $\textmd{BERT}_{fast}$  and $\textmd{BERT}_{ori}$ models trained for the same training time.
We train a $\textmd{BERT}_{fast}$ model for totally 400,000 training steps, and among them 100,000 steps are used for the first phase.  The baseline is the $\textmd{BERT}_{ori}$ model trained for 400,000 steps.
We  show  the curves of the F1 scores with respect to the training time in Figure \ref{fig:1}. We can see that, for the same training time, $\textmd{BERT}_{fast}$  always performs better than $\textmd{BERT}_{ori}$. Note that $\textmd{BERT}_{fast}$ trained for about 15 hours achieves better performance than $\textmd{BERT}_{ori}$ trained for 25 hours (reduce about $40\%$ training time), which demonstrate the effectiveness and efficiency of the proposed method.

\subsection{Evaluation on GLUE}
The General Language Understanding Evaluation (GLUE \cite{wang2018glue}) benchmark  consists of nine natural language understanding tasks,  namely Corpus of Linguistic Acceptability (CoLA \cite{CoLA}), Microsoft Research Paraphrase Corpus (MRPC \cite{MRPC}), Semantic Textual Similarity Benchmark (STS \cite{STS}), Stanford Sentiment Treebank (SST \cite{SST}), Multi-Genre NLI (MNLI \cite{MNLI}),  Quora Question Pairs (QQP \cite{QQP}), Question NLI (QNLI \cite{QNLI}), Recognizing Textual Entailment (RTE \cite{RTE}), and Winograd NLI (WNLI \cite{WNLI}). Following \cite{BERT}, we also exclude the problematic WNLI  task. For each task, we fine-tune the pre-trained model for 3 epochs with batch size 32, and  we perform a grid search on the learning rate set  [5e-5, 4e-5, 3e-5,  2e-5]. 
Table \ref*{tb:3} shows the median of development set  results over five different initializations.  Further, we  report the online test results  of the BERT-Base  models pre-trained for 400,000 steps, which are shown in Table \ref*{tb:4}. Both the development and test results show that with the same training steps, our method can achieve similar performance to the original BERT, while  the training time is greatly reduced. The results are in line with those shown in Subsection \ref{sec:5-2-1}.

\begin{table*}[t]\centering 
	\caption{Dev set results of  GLUE tasks (except WNLI) for the pre-trained models.  F1 scores are reported for QQP, MRPC, and Spearman correlations are reported for STS-B. The accuracy scores are reported for the other tasks.}
	\label{tb:3}
	\hspace*{\fill} \\
	\begin{tabular}{ |c| c| c| c| c |c |c| c|c |c |c|}
		\hline	{Model} &Steps&MNLI&QQP&QNLI&SST-2 & CoLA&STS-B&MRPC&RTE&Avg \\ 
		\hline
		$\textmd{BERT}_{ori}$&100,000&81.9/81.6&87.1 &89.2&90.9&49.9& 87.9&89.2&66.4 &80.5\\	
		\hline
		$\textmd{BERT}_{fast}$&100,000&81.5/81.2&87.2&88.9&90.7&49.2& 87.5&89.5&65.7& 80.2\\			
		\hline
		\hline
		$\textmd{BERT}_{ori}$&200,000&82.8/83.3&87.7 &89.8&90.6&51.7& 88.6&89.5&65.3&81.0\\	
		\hline
		$\textmd{BERT}_{fast}$&200,000&82.8/83.1&87.6 &89.9&90.3&51.3& 88.4&89.4&67.5& 81.1\\			
		\hline
		\hline
		$\textmd{BERT}_{ori}$&300,000&83.4/83.8&88.2 &90.8&91.2&56.5&88.9&89.8&65.3&82.0\\	
		\hline
		$\textmd{BERT}_{fast}$&300,000&83.2/83.5&87.9 &90.2&90.9&56.0& 88.7&89.6&66.2& 81.8\\			
		\hline
		\hline
		$\textmd{BERT}_{ori}$&400,000&83.8/84.2&88.5 &91.1&91.6&57.1& 89.2&90.4&68.1  &82.7\\	
		\hline
		$\textmd{BERT}_{fast}$&400,000&83.6/83.9&88.4 &90.8&91.4&56.8& 89.3&90.1&67.5& 82.4\\			
		\hline
	\end{tabular}
\end{table*}

\begin{table*}[t]\centering 
	\caption{Online test results of GLUE tasks. }
	\label{tb:4}
	\hspace*{\fill} \\
	\begin{tabular}{ |c| c| c| c |c |c| c|c |c |c|}
		\hline	
		{Model} &MNLI&QQP&QNLI&SST-2 & CoLA&STS-B&MRPC&RTE&Avg \\ 
		\hline
		$\textmd{BERT}_{ori}$&84.2/82.9&71.4&90.2&92.9&53.9&84.6& 88.3&65.5 &79.3\\	
		\hline
		$\textmd{BERT}_{fast}$&83.9/82.7&71.0 &90.0&92.6&52.8& 84.2&88.5&64.9& 79.0\\
		\hline
	\end{tabular}
\end{table*}
\subsection{Combined with ALBERT and Progressively Stacking}
As stated before, our method can be well combined with the existing training speedup methods, including ALBERT \cite{ALBERT} and progressively stacking \cite{Stacking}. Table \ref{tb:5} shows the performance in terms of accuracy and speed. All the models adopt the BERT-Base hyperparameter setting. We training each model for totally 400,000 step. For ALBERT, we first train a relaxed ALBERT model for 200,000 steps and use the remaining steps for refined training. In \cite{Stacking}, progressively stacking trains a BERT-Base model by first training a 3-layer BERT for 50,000 steps. Then the 3-layer model is   stacked into a 6-layer BERT, which is trained for 70,000 steps. Finally,  the 6-layer BERT is stacked into a 12-layer BERT, and  the 12-layer BERT is trained for 280,000 steps. Combining with our method and progressively stacking, we first train a  3-layer relaxed BERT for 50,000 steps. Then we stack it into a 6-layer relaxed BERT and train the 6-layer model for 70,000 steps. Next, we further stick the model into a relaxed 12-layer BERT and train the relaxed 12-layer model for 80,000 steps. Finally, we restore the original BERT-Base model from the trained relaxed 12-layer BERT, and retrain the model for the remaining 200,000 steps. We also train the models using the original ALBERT and progressively stacking methods as the baselines.
\begin{table*}[t]\centering 
	\caption{Online test results of GLUE tasks. }
	\label{tb:5}
	\hspace*{\fill} \\
	\begin{tabular}{ |c| c| c| c |c |c| c|c |c |c|c|}
		\hline	
		{Model} &Total step&Time&MNLI&QNLI&SST-2 &STS-B&MRPC&SQuAD&Avg &Speedup\\ 
		\hline
		ALBERT&400,000&22.25 h&81.4/82.1& 89.2&89.6& 87.8& 87.4&87.2& 86.4&1.0$\times$\\	
		\hline
		ALBERT+CoRe&400,000&17.42 h&81.1/82.0  &89.4&89.3&87.9&87.3&87.4&86.3 &1.3$\times$\\
		\hline
		Stacking&400,000&20.12 h&83.6/84.1 &91.2&91.8 & 88.4&89.9&89.5& 88.4 &1.1$\times$\\
		\hline
		Stacking+CoRe&400,000&17.01 h&83.4/83.9 &91.0&91.5 & 88.3&89.5&89.7&88.2 &$1.3\times$\\
		\hline
	\end{tabular}
\end{table*}
From Table \ref{tb:5}, we can see that for both ALBERT and progressively stacking, when they are used in conjunction with our method, the training speed can be further improved without performance degradation.

\subsection{Ablation Study}
\begin{figure}[t]
	\centering
	\includegraphics[scale=0.6]{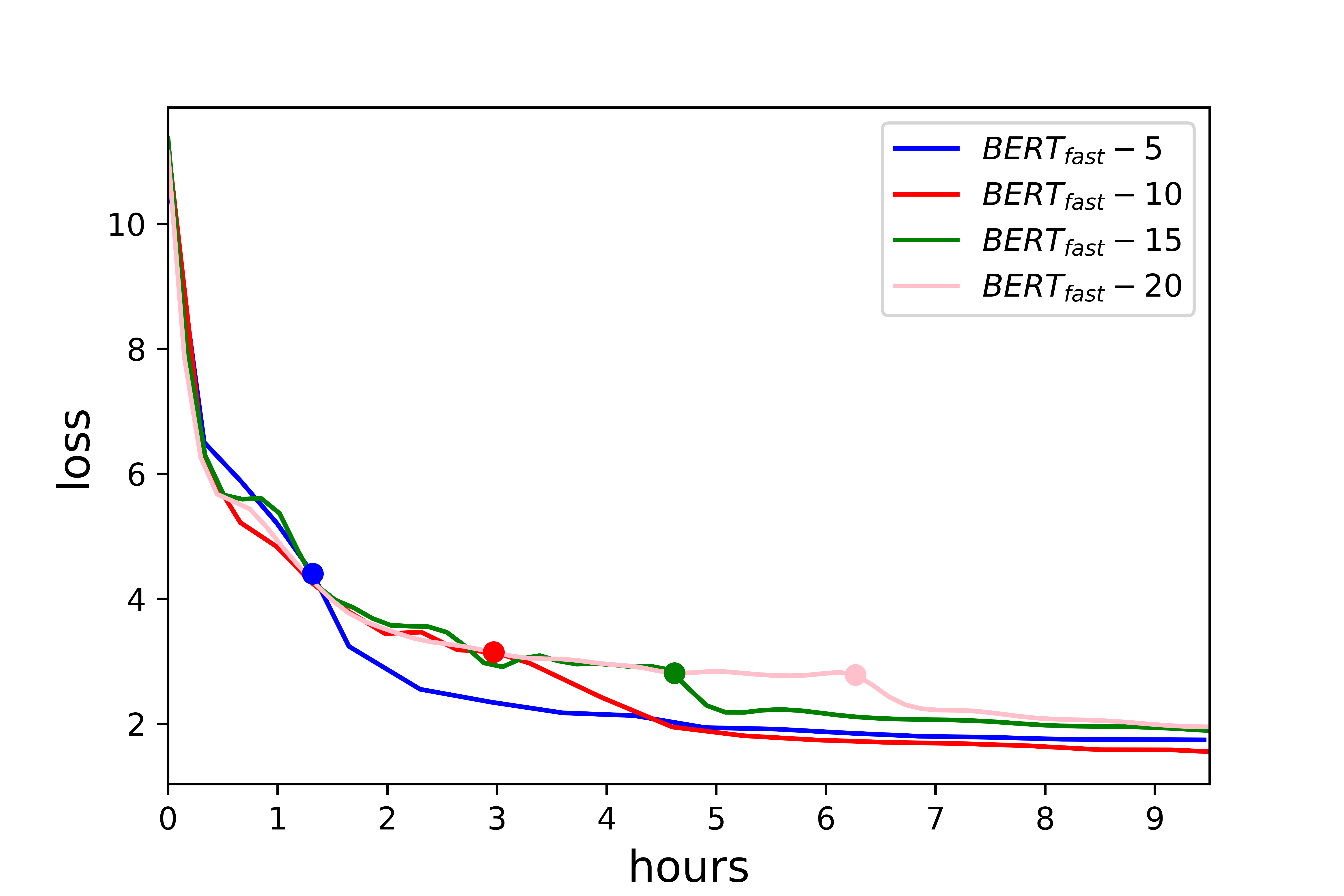}
	\caption{\label{fig:3} This figure shows the  loss  curves of $\textmd{BERT}_{fast}$ models with various training time in the first phase. The x-axis is the wall time of training.}
\end{figure}

From Subsection \ref{sec:5-2-1}, we can see that with totally 400,000 training steps, the $\textmd{BERT}_{fast}$ model with 100,000 steps used for the first phase performs better than that with 200,000 steps used for the first phase (89.2955 F1 score vs 88.8421 F1 score on SQuAD v1.1). However,  the former consumes more training time than the latter (22.36 hours vs 19.43 hours). So in limited training time, the allocation of the time for coarse training and refined training  is  a key factor that determines the performance of the model. In this subsection, we investigate the effect of the allocation proportion. For clarity, we  denote the models trained with 50000, 100000, 150000, and 200000 steps in the first phase  by $\textmd{BERT}_{fast}-5$, $\textmd{BERT}_{fast}-10$, $\textmd{BERT}_{fast}-15$, and $\textmd{BERT}_{fast}-20$, respectively. We first show the pre-training loss curves  of these $\textmd{BERT}_{fast}$ models in Figure \ref{fig:3} (for clarity, only a part of loss is presented). From the figure, we see that for each  $\textmd{BERT}_{fast}$ model,  its loss curve has  an inflection point (marked by a small circle), which corresponds to point-in-time that the training process switches from the coarse pre-training phase to the refined retraining phase.  Note that after the  inflection point the loss does not increase significantly but rapidly decreases, which demonstrates that in the retraining phase, the recovered original BERT model inherits the performance of the relaxed model. In addition, we see that the loss curves of $\textmd{BERT}_{fast}-5$ and $\textmd{BERT}_{fast}-10$  are close and are lower than those of $\textmd{BERT}_{fast}-15$ and $\textmd{BERT}_{fast}-20$ in most instances. That is because the relaxed BERT model is much more simple than the original BERT, and it can only learn some coarse information from the data. After a period of  time, the   relaxed model quickly achieves convergence and further training cannot bring performance improvement. So it is not a good choice to spend too much time in the first phase  training the relaxed BERT model.  We should switch the training mode to the second phase when the loss does not decrease significantly for a period of time. 

\begin{figure}[t]  
	\centering
	\includegraphics[scale=0.6]{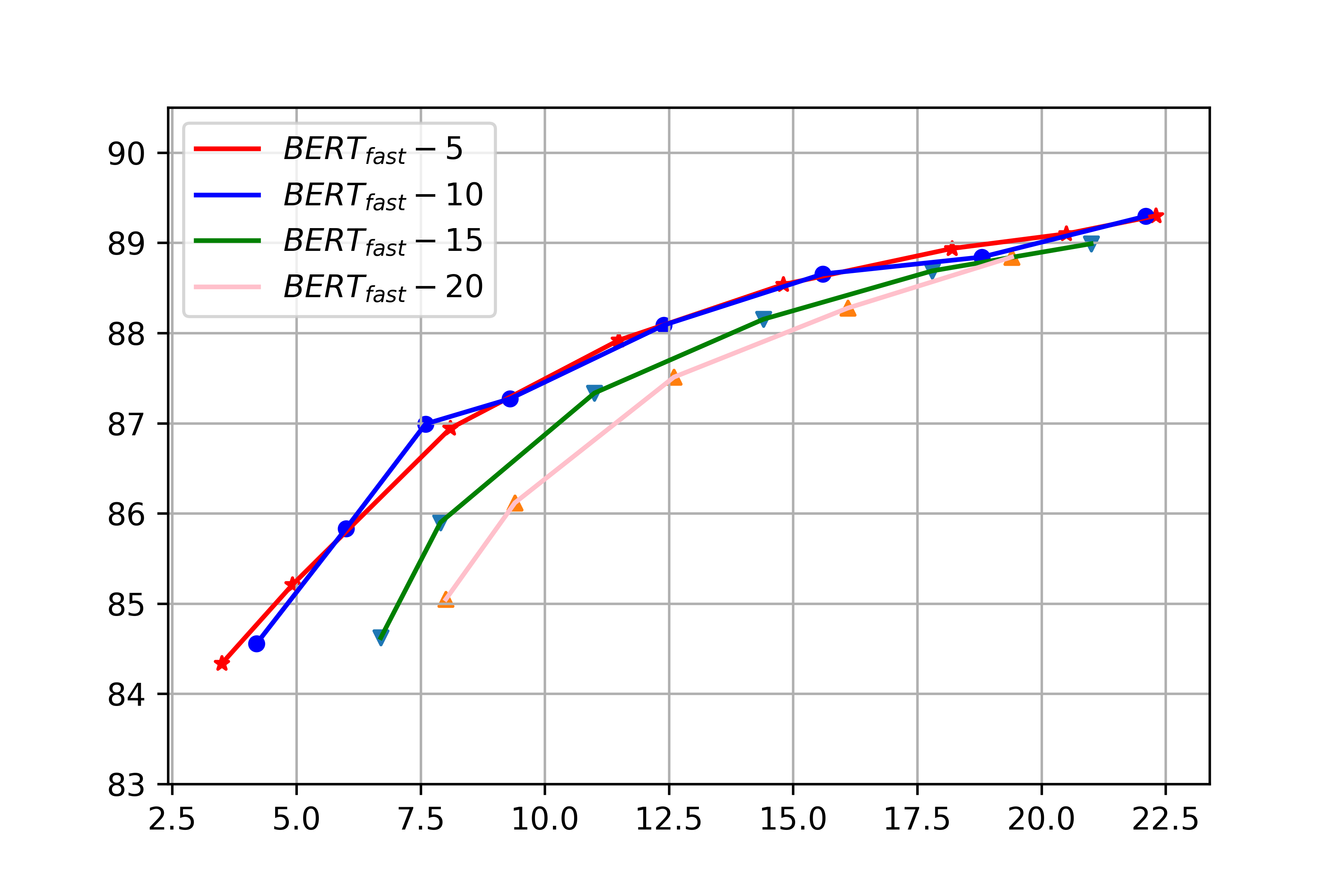}
	\caption{\label{fig:2} The F1 score curves of  $\textmd{BERT}_{fast}$ models on SQuAD v1.1 with different training steps used in the first phase. The total steps are 400,000. The x-axis denotes the wall time of training.}
\end{figure}

Further, we  show the  F1 scores on SQuAD v1.1 of these $\textmd{BERT}_{fast}$ models in Figure \ref{fig:2}. For all the models, we only show the results after the first phase is finished, since the relaxed model is not used for dealing with downstream tasks.
The results shown in Figure \ref{fig:2} is in line with those shown in Figure \ref{fig:3}. The performance of $\textmd{BERT}_{fast}-5$ and $\textmd{BERT}_{fast}-10$ is similar and is always better than the performance of  $\textmd{BERT}_{fast}-15$ and $\textmd{BERT}_{fast}-20$. 

\section{Conclusion}
In this paper,  we proposed an  efficient method named CoRe, to speed up the training of BERT. We decomposed the training process into two phases. In the first phase, we constructed a relaxed substitute for the original BERT model. The relaxed model has much lower model complexity so that it can be quickly trained.  When the first phase is finished, we transfer its knowledge to an original standard BERT model, and in the second phase, we further retrain the original  BERT on this basis.  Due to the good initialization provided by the relaxed model, the training process is greatly accelerated.   Experimental results showed that compared with the original method that trains the model from scratch with a random initialization, the proposed CoRe method can achieve competitive performance, but consumes much less training time. In addition, CoRe method can be  used in conjunction with the other training speed up techniques to achieve higher training efficiency.

\bibliographystyle{named}
\bibliography{sample-base}


\end{document}